\def\year{2019}\relax

\documentclass[letterpaper]{article} 
\usepackage{aaai19}  
\usepackage{times}  
\usepackage{helvet}  
\usepackage{courier}  
\usepackage{url}  
\usepackage{graphicx}  
\usepackage{booktabs}       
\usepackage{amsmath,amssymb}
\usepackage{microtype}
\usepackage{bbm}
\usepackage{bm}
\DeclareMathOperator*{\argmax}{argmax}
\newcommand\smalleq{\mkern1.5mu{=}\mkern1.5mu}

\newcommand{\R}{\mathbb{R}}
\newcommand{\norm}[1]{\|#1\|}

\newcommand{\corelL}{\mathcal{L}_{\textit{COREL}}}
\newcommand{\arL}{\mathcal{L}_{\textit{AR}}}
\newcommand{\cceL}{\mathcal{L}_{\textit{CCE}}}
\newcommand{\centerL}{\mathcal{L}_{\textit{center}}}
\newcommand{\attL}{\mathcal{L}_{\textit{attract}}}
\newcommand{\repL}{\mathcal{L}_{\textit{repulse}}}

\newcommand{\X}{\mathcal{X}}
\newcommand{\y}{\mathcal{Y}}
\newcommand{\h}{\mathcal{H}}

\newcommand{\rep}{F_{\theta}}
\newcommand{\pred}{G_{\textbf{W}}}
\newcommand{\sumK}{\sum_{k=1}^{K}}

\newcommand{\hi}{\textbf{h}^{(i)}}
\newcommand{\wk}{\textbf{w}_{k}}
\newcommand{\wyi}{\textbf{w}_{y_i}}
\newcommand{\W}{\textbf{W}}

\newcommand\blfootnote[1]{%
  \begingroup
  \renewcommand\thefootnote{}\footnote{#1}%
  \addtocounter{footnote}{-1}%
  \endgroup
}

\begin{document}
%
\title{Clustering-Oriented Representation Learning with Attractive-Repulsive Loss}
\author{
Kian Kenyon-Dean$^{*,\dagger}$ \and Andre Cianflone$^{*,\dagger}$ \and Lucas Page-Caccia$^{*,\dagger}$\\
\{\url{first.last}\}\url{@mail.mcgill.ca}
\AND Guillaume Rabusseau$^{*,\ddagger}$\\
\url{guillaume.rabusseau@umontreal.ca}
\And Jackie Chi Kit Cheung$^{*,\dagger}$\\
\url{jcheung@cs.mcgill.ca}\\
\And Doina Precup$^{*,\dagger}$\\
\url{dprecup@cs.mcgill.ca}
}
\maketitle


\begin{abstract}
The standard loss function used to train neural network classifiers, categorical cross-entropy (CCE), seeks to maximize accuracy on the training data; building useful representations is not a necessary byproduct of this objective. 
In this work, we propose \emph{clustering-oriented representation learning} (COREL) as an alternative to CCE in the context of a generalized \emph{attractive-repulsive} loss framework. 
COREL has the consequence of building latent representations that collectively exhibit the quality of \emph{natural clustering} within the latent space of the final hidden layer, according to a predefined similarity function.
Despite being simple to implement, COREL variants outperform or perform equivalently to CCE in a variety of scenarios, including image and news article classification using both feed-forward and convolutional neural networks.  
Analysis of the latent spaces created with different similarity functions facilitates insights on the different use cases COREL variants can satisfy, 
where the \textit{Cosine}-COREL variant makes a consistently clusterable latent space, while \textit{Gaussian}-COREL consistently obtains better classification accuracy than CCE.
\end{abstract}

\section{Introduction}
\blfootnote{$*$: Mila Qu\'{e}bec AI Institute. $\dagger$: Department of Computer Science, McGill University. $\ddagger$ Department of Computer Science and Operations Research, Universit\'{e} de Montr\'{e}al. Qu\'{e}bec.}%
The last hidden layer of a neural network most powerfully expresses the network's ``understanding'' of the input with respect to the its training objective \cite{zeiler2014visualizing}. 
Neural classifiers are typically trained using the categorical cross-entropy loss function (CCE), which induces the network to learn how to project the input data into a linearly separable latent space. 
We argue that the focus of the training process can be beneficially shifted from optimizing for high classification accuracy to optimizing for \emph{structured} latent representations in the final layer of a neural model. 
More precisely, beyond linear separability in the latent space, the latent representations should be \emph{naturally clusterable}~\cite{bengio2013representation}: samples belonging to the same class should be concentrated within a distinct cluster in the latent space, and clusters corresponding to different classes should be distinguished by low density regions. 
This is a generic prior for real-world data, and models that reflect this prior are likely to have better generalization performance~\cite{bengio2013representation}. 

We propose \textit{Clustering-Oriented Representation Learning} (COREL) as a novel perspective for designing neural networks. COREL views classification as a problem of building naturally clusterable latent representations of data, where clusterability is defined with respect to some similarity function. We present COREL within our proposed \textit{attractive-repulsive} loss framework. This framework generalizes CCE as a specific case of attractive-repulsive loss that uses the dot product as the similarity function. Our general framework offers insights into using alternative similarity functions that are oriented toward natural clusterability.

We experiment with two similarity functions, cosine similarity and Gaussian similarity, each with specific formulations for attractive and repulsive signaling. Experiments on image classification and document-topic classification reveal that these COREL variants create naturally clusterable latent spaces, unlike CCE. We find that basic clustering algorithms applied to the COREL latent spaces can obtain test set accuracy that matches the accuracy earned when the learned output weight vectors are used for inference. While the \textit{Cosine}-COREL variant makes a consistently clusterable latent space, we find that \textit{Gaussian}-COREL creates representations that are consistently better in accuracy than both CCE and center loss in almost every classification scenario. 

This work proceeds as follows: we first present attractive-repulsive loss. Next, we present clustering-oriented representation learning and our two proposed variants. We then present the specific details of our experimental design. Next, we report classification performance on the three datasets and analyze the clusterability of the representations obtained by different model variants. Finally, we conclude with the related work and future perspectives.\footnote{All code and resources used in this work can be found at:\\ \url{https://github.com/kiankd/corel2019}}

\section{Attractive-Repulsive Loss}
We propose \textit{Attractive-Repulsive loss} (AR) as a new general class of optimization objectives for understanding loss functions and inference in neural models. AR loss is defined with respect to an input space $\X$, latent space $\h$, and output space $\y$; we divide the feed-forward process of neural networks into two steps, a \textit{representation} mapping $\rep: \X \rightarrow \h$, followed by a \textit{prediction} mapping $\pred: \h \rightarrow \y$. 

\textbf{Representation.} This is a function parameterized by $\theta$, $\rep(x) = \textbf{h}$. Via a composition of nonlinear transformations (such as those defined by a CNN), it projects input samples $x$ to  $\h$, the $H$-dimensional latent space represented by the \textit{final hidden layer} of the model, where $\textbf{h} \in \R^{H}$.
    
\textbf{Prediction.} The network output function $\pred(\textbf{h}) = \textbf{y}$, a vector of values indicating model confidence of membership to each class. $\pred$ maps $\textbf{h}$ to output space $\y$ using a matrix\footnote{Without loss of generality, we omit the bias vector of the output function as it can be incorporated with a component of $\textbf{h}$.} $\W \in \R^{K \times H}$, with a row for each of the $K$ classes. In the AR perspective, the prediction function measures the \textit{similarity} of a sample to the representation of a class $\wk$. Therefore, the $k^{\text{th}}$ component of the prediction vector is defined by a similarity function $\pred(\textbf{h})_{k} = s(\textbf{h}, \wk)$, and thus the final class prediction is the most similar class to the sample: $\argmax_{k} s(\textbf{h}, \wk)$.

Given the definitions of representation and prediction, the fundamental components of the AR loss are: (1) the similarity function, $s$; and, (2) the definition of an \textit{attractive} loss ($\attL$), and a \textit{repulsive} loss ($\repL$). The full loss generalization is defined as:
\begin{equation} \label{eq:arl}
\begin{split}
\arL = \sum_{i=1}^{N} \bigg[ &-\attL^{s}(\hi, \wyi) \lambda \\
&+ \repL^{s}(\hi, \W)(1-\lambda) \bigg]
\end{split}
\end{equation}
where the attractive and repulsive loss functions are defined with respect to the given similarity function $s$ and the hyperparameter $\lambda \in (0,1]$, which mediates between the two loss terms. Note that if $\lambda\smalleq 0$ it is unlikely that there can be a solution to the optimization. 

The attractive loss ($\attL$) deals solely with the weight vector corresponding to the class to which sample $i$ belongs, $\wyi$. Its purpose is to make samples and weight vectors for their classes as similar as possible by being trained with the gradient descent objective of minimizing their dissimilarity. Meanwhile, the repulsive loss ($\repL$) can use the output weight matrix $\W$ in its entirety in order to minimize the similarity of samples to the other classes, and possibly to work as a normalization factor, as in CCE. Defining these components of $\arL$ requires careful consideration of the similarity function and what properties would be desirable to impose within the latent space.

\subsection{Categorical Cross-Entropy} \label{sec:cce}
Categorical cross-entropy (CCE) is the standard loss function used to train neural networks to solve classification tasks. CCE's final objective is to learn a linear transformation defined by $\W$ such that all of the training data projected onto $\h$ becomes linearly separable into the $K$ classes. While CCE seeks to learn a linear separation on $\h$, $\h$ is a nonlinear space resulting from $\rep: \X \rightarrow \h$;  this linear separability is thus nonlinear with respect to the original input space $\X$.

CCE is traditionally understood as modeling the probability that a sample \textit{i} belongs to class \textit{k}. While such a perspective offers useful theoretical intuitions, it does not offer always substantial utility in practice, as the posterior probabilities predicted using softmax can be overly confident, especially for adversarial examples \cite{wan2018rethinking}. The CCE loss function seeks to maximize the log-likelihood of the $N$-sample training set, where $y_i$ is the index of the true class for sample $i$:
\begin{equation*}
\cceL = - \sum_{i=1}^N \log \frac{\exp(\wyi^{T}\hi)}{\sumK \exp(\wk^{T}\hi)} 
\end{equation*}

With simple algebraic manipulation, the following reformulation of the CCE loss function lends itself to the AR expression of CCE,
\begin{equation} \label{eq:corelcce}
	\cceL = \sum_{i=1}^{N}  - s(\hi,\wyi) + \log \sumK e^{s(\hi, \wk)}
\end{equation}
where the similarity function is $s(\hi, \wyi) = \wyi^{T}\hi$. This formulation of CCE reveals the \textit{attractive} component in the first half of the equation. By being trained to minimize $\cceL$, the model will seek to maximize the inner product -- the similarity -- between a sample's representation and the vector corresponding to its class. The second term above can be viewed as the {\em repulsive} component, which should be minimized. Note that this term is more complex. Interestingly, it also includes the attractive objective, which has traditionally been understood as necessary for a consistent probabilistic output. Our perspective offers an alternative insight: CCE gives contradictory signals to the model, which is necessary in order to prevent the divergence of the weight vectors caused by maximizing the inner product, as this is an unbounded similarity function. 

This analysis reveals that the CCE objective creates a latent space dependent on the inner product. CCE learns weight vectors that separate the representations into their classes. We hypothesize that bounded similarity functions can be used to learn weight vectors that generally represent the class itself; e.g., by being the center of representations belonging to that class, rather than as an intermediate separator between classes. 
With either different definitions of vector similarity, different forms of attractive and repulsive losses, or both, one can induce different desirable qualities into the latent space. 
This observation motivates our pursuit to design loss functions that produce naturally clusterable latent spaces.

\section{Clustering-Oriented Representation Learning} \label{sec:corel}

Clustering-oriented representation learning (COREL) is based on learning latent representations of data that capture the quality of \textit{natural clustering} inherent in real-world data \cite{bengio2013representation}. 
Thus, COREL loss functions are aimed toward building representations that are easily clusterable into their corresponding classes within the latent space, according to a similarity function. 

The basic insights of clustering algorithms inform COREL: samples should be \textit{similar} to other samples belonging to the same class, and \textit{dissimilar} from samples belonging to other classes. Therefore, the two fundamental components for making a clusterable latent space are \textit{attraction} and \textit{repulsion}, as expressed by the generic AR loss function in Equation~\ref{eq:arl}. 

\subsection{Similarity Functions}
While CCE defines one specific similarity function, the inner product, there are many alternative functions that could be used, each of which will offer different qualities in the latent space. In this work, we experiment with two similarity functions: \textit{cosine similarity} and \textit{Gaussian similarity}.

\subsubsection{Cosine similarity.} The cosine similarity between two vectors measures the cosine of the angle between them. The definition $s_{cos}$ is,
\begin{equation} \label{eq:scos}
s_{cos}(\textbf{h}, \textbf{w}) = \frac{\textbf{h}}{\norm{\textbf{h}}} \cdot \frac{\textbf{w}}{\norm{\textbf{w}}} 
\end{equation}
which is bounded between $-1$ and $1$. When the cosine similarity between two vectors is zero they become orthogonal -- this will prove important later in how we define the attractive and repulsive losses with respect to this similarity function. 

\subsubsection{Gaussian similarity.}
We define Gaussian similarity based on the univariate normal probability density function and the standard RBF kernel function. A hyperparameter $\gamma = \frac{1}{2\sigma^{2}}$ is introduced, thus defining Gaussian similarity $s_{gau}$ as,
\begin{equation} \label{eq:sgaus}
s_{gau}(\textbf{h}, \textbf{w}) = -\gamma \norm{\textbf{h} - \textbf{w}}^{2}
\end{equation}
which is bounded between $-\infty$ and $0$. This similarity function has the property of being the traditional squared error function when $\gamma=0.5$, the importance of which is highlighted in the next section.

\begin{figure*}[t]
\centering
\begin{minipage}[b]{.4\textwidth}
\centering
\includegraphics[width=\linewidth]{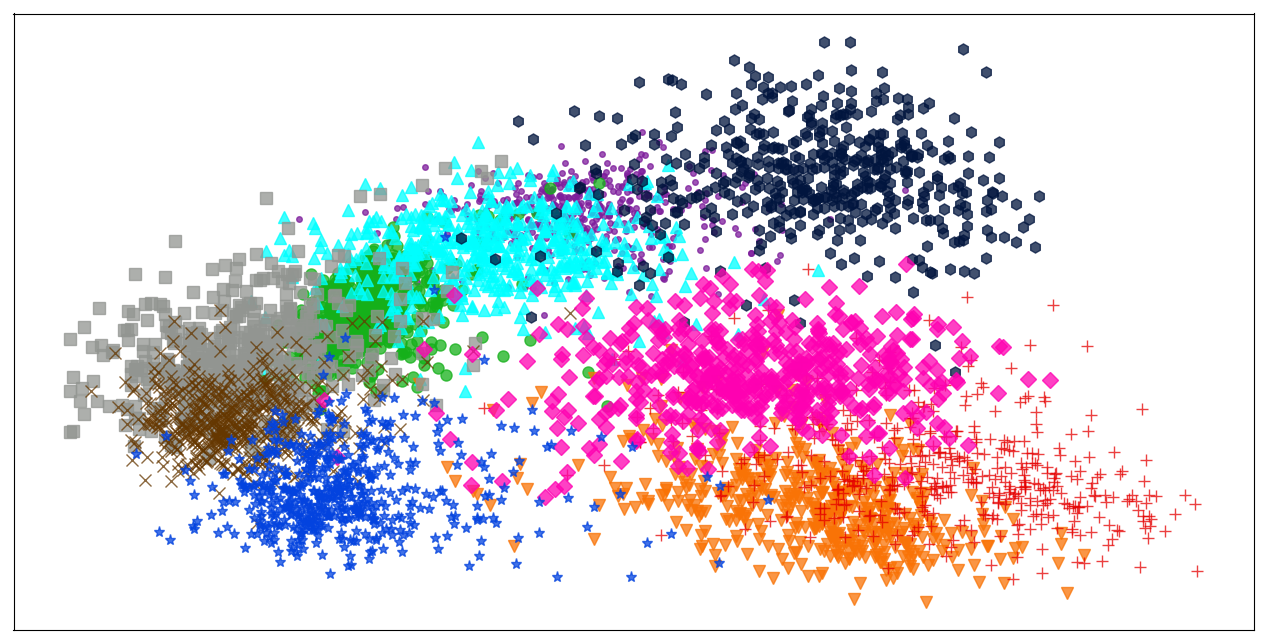}
\end{minipage}\qquad
\begin{minipage}[b]{.4\textwidth}
\centering
\includegraphics[width=\linewidth]{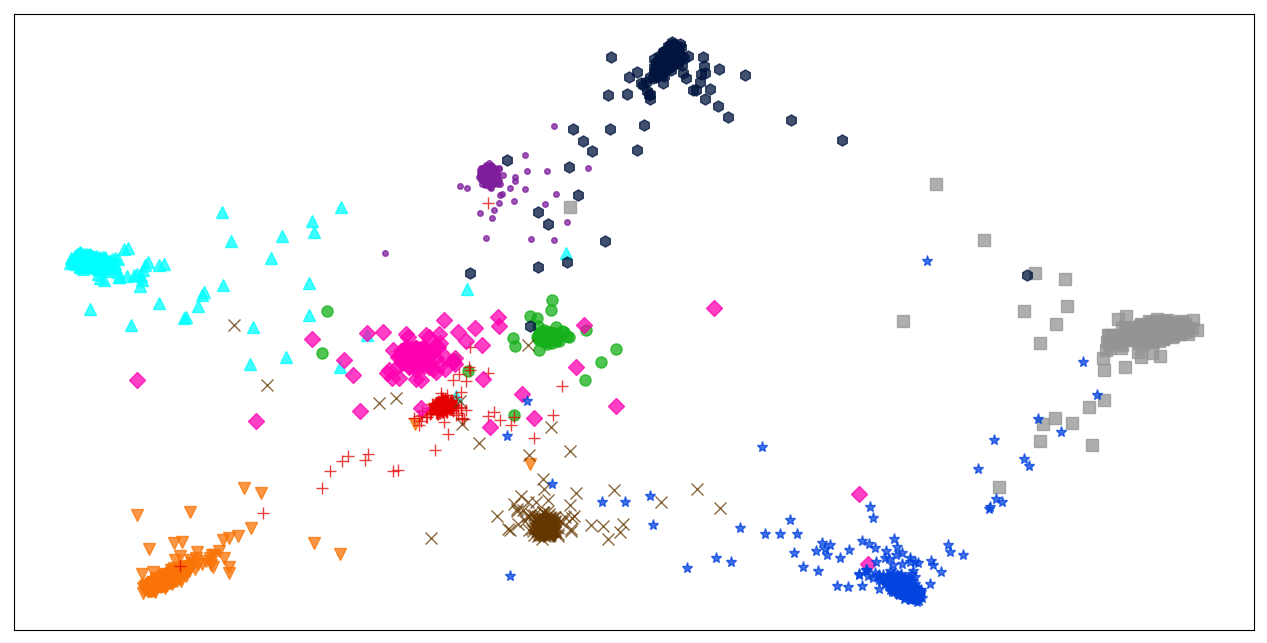}
\end{minipage}
\begin{minipage}[b]{.4\textwidth}
\centering
\includegraphics[width=\linewidth]{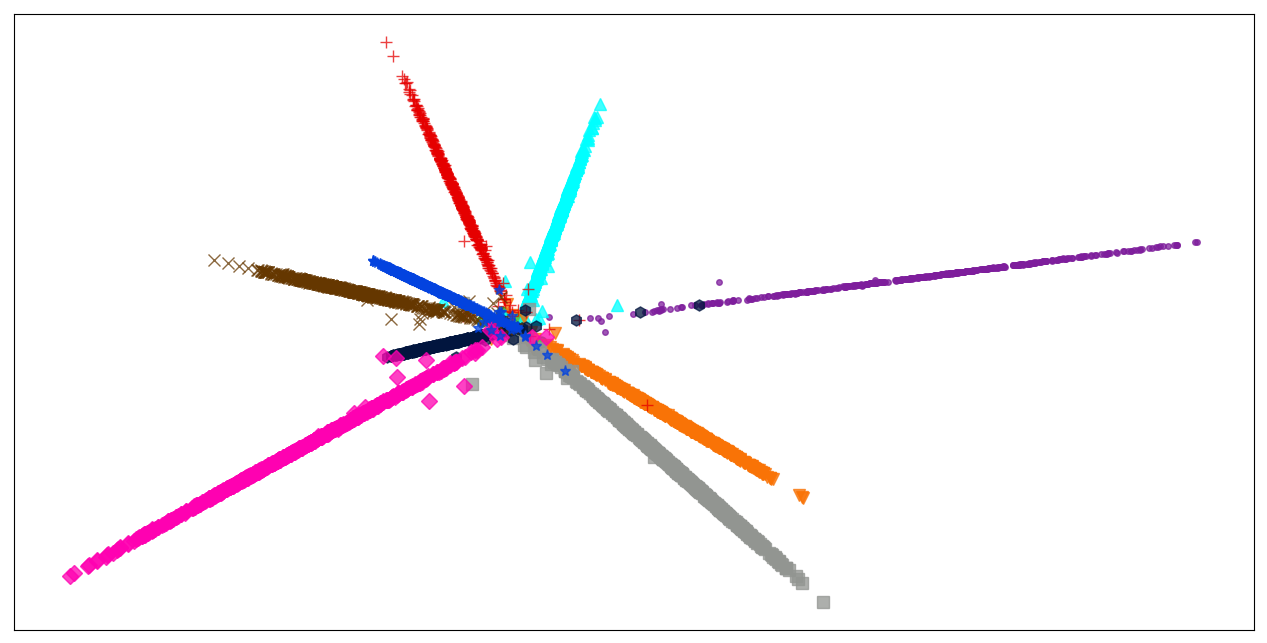}
\end{minipage}\qquad
\begin{minipage}[b]{.4\textwidth}
\centering
\includegraphics[width=\linewidth]{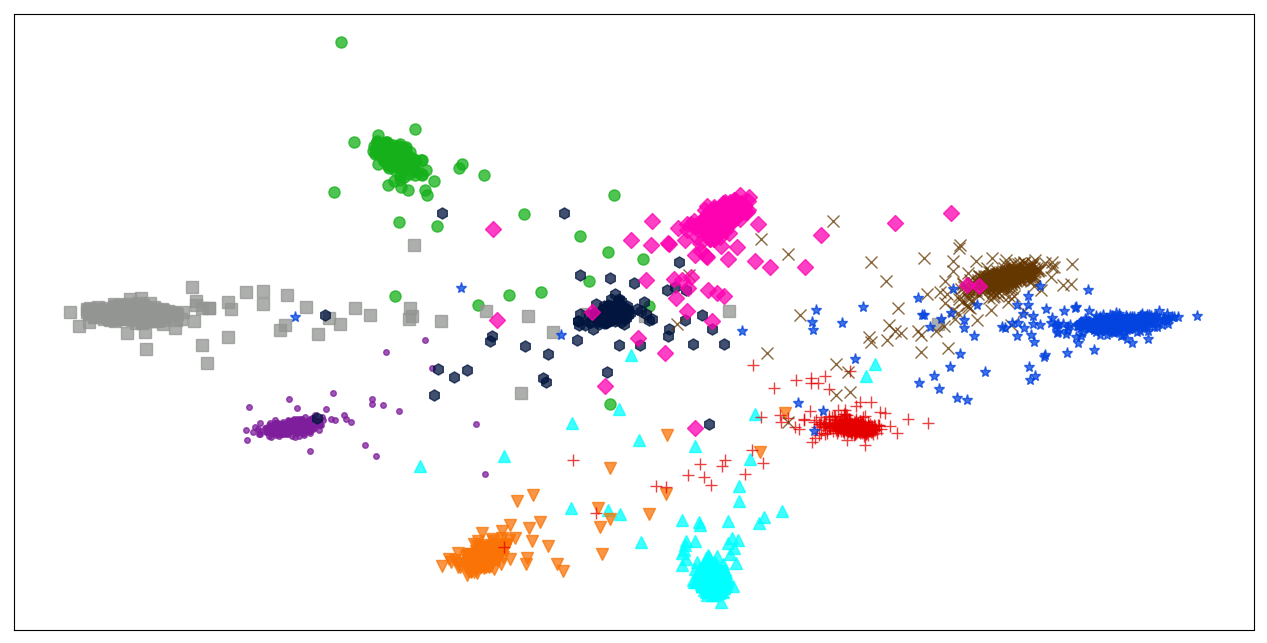}
\end{minipage}
\begin{minipage}[b]{\textwidth}
\centering
\includegraphics[width=0.8\linewidth,trim={0cm 5.75cm 0cm 5.85cm},clip]{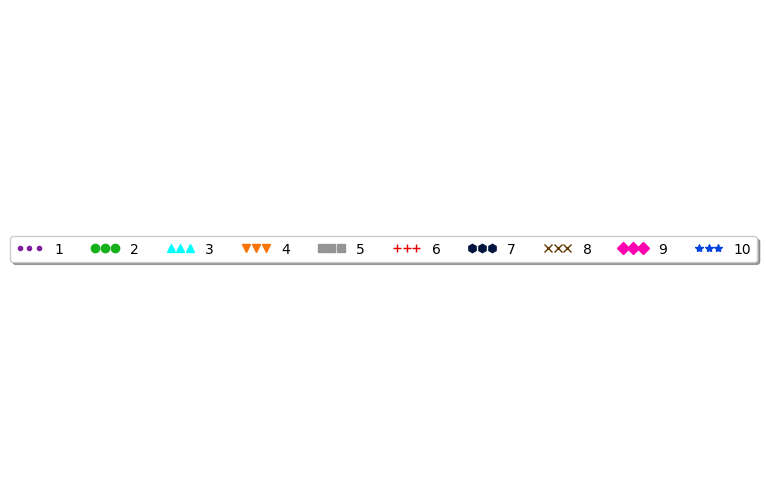}
\end{minipage}
\caption{Visualization of MNIST test set representations made with a CNN, compressed from $128$ dimensions with PCA. Top left is CCE, top right is center loss, bottom left is \textit{Cosine}-COREL, bottom right is \textit{Gaussian}-COREL.}
\label{fig:viz}
\end{figure*}

\subsection{Attractive-Repulsive COREL Loss}
Attractive and repulsive losses are defined in accordance with the geometric features and bounds of the similarity function. Here, we propose different formulations for each for the two similarity functions.

\subsubsection{Cosine similarity loss.}
We define the following attractive and repulsive losses for cosine similarity (Equation~\ref{eq:scos}):
\begin{equation} \label{eq:lcos}
\begin{split}
&\attL^{cos} = s_{cos}(\hi, \wyi)\\
&\repL^{cos} = \max_{k\neq y_i} s_{cos}(\hi, \wyi)^{2}
\end{split}
\end{equation}
The definition of attractive loss is to maximize the cosine similarity between a sample and the weight vector for its class. However, the fundamental insights here are found in the definition of repulsive loss, $\repL^{cos}$. First, by using the $\max$ operation (one may call it \textit{hardmax}) a very explicit objective is expressed: to minimize the cosine similarity between a sample and the weight vector of the class to which it is most similar \textit{but to which it does not belong}. The second feature of this loss is squaring the cosine similarity. This is necessary in order to have the model optimize for \textit{orthogonal} classes in the latent space, thus defining an orthogonal basis. If it were not squared, the model could optimize toward creating vectors that are co-linear (at a $180$-degree angle with each other), which is intuitively not desirable\footnote{Moreover, preliminary experiments revealed that squaring the similarity in $\repL^{cos}$ improves generalization better than when either not doing so or taking the absolute value.}. 

The softmax objective is problematic when applied to cosine similarity, the normalized counterpart to the inner product. The upper bound on the maximum probability that could be assigned to a specific sample, when using cosine similarity in the softmax, is $\frac{e^2}{e^2 + K - 1}$, which means that the maximal probability is only $7\%$ when $K\smalleq100$, for example. This has also been observed by \cite{qi2017learning} and \cite{wang2017normface}; both introduce a learnable scaling factor to resolve this issue. On the other hand, our proposed cosine similarity-based loss (Equation~\ref{eq:lcos}) requires neither a probabilistically-informed motivation nor an extra parameter in order to be well-behaved. 

\subsubsection{Gaussian similarity loss.}
Gaussian similarity (Equation~\ref{eq:sgaus}) is amenable to the softmax formulation in CCE because the maximal dissimilarity is unbounded. Additionally, the exponential function included in the softmax operation induces the Gaussian quality of the similarity function into the latent space. Intuitively, it makes sense to maintain $s_{gau}(\hi, \wyi)$ in $\repL^{gau}$; otherwise, the model could be optimized by making the representations of samples of every class completely the same as their class's weight vector. We define the \textit{softmax-informed} loss for Gaussian similarity as:
\begin{equation} \label{eq:lgaus}
\begin{split}
&\attL^{gau} = s_{gau}(\hi, \wyi)\\
&\repL^{gau} = \log \sum_k e^{s_{gau}(\hi, \wk)}
\end{split}
\end{equation}
This formulation reveals the relevance of the Gaussian interpretation of the similarity function. By using the softmax operation in the loss function, the model is really maximizing a Gaussian probability density function with a diagonal covariance matrix $\Sigma$ with entries $\sigma^2$, due to the equality: 
\begin{equation*}
	\exp-\frac{1}{2\sigma^2} \norm{\textbf{h} - \textbf{w}}^2 = \exp -\frac{1}{2} (\textbf{h} - \textbf{w})^\intercal \Sigma^{-1} (\textbf{h}-\textbf{w})
\end{equation*}

In the present work, we are motivated to make clustering-oriented latent spaces with simple-to-implement loss functions. We thus assume, for this loss function, that the model can construct a latent space such that the classes can be expressed by a simple univariate Gaussian, keeping $\gamma=0.5$ for the experiments in this work.

\subsection{Relation to K-Means}
Further examination of $\attL^{gau}$ reveals the relationship between the Gaussian-similarity loss and the K-Means objective. When $\gamma=0.5$, $\attL^{gau}$ defines the mean squared error. Referring back to the general AR loss in Equation~\ref{eq:arl}, the maximum-likelihood solution of using the mean squared error for classification (that is, for the ``nearest-centroid'' classifier \cite{schutze2008introduction}) emerges in a closed-form when only the attractive signal is considered in the loss (i.e., when $\lambda=1$):
\begin{equation}
\begin{split}
\corelL^{gau} (\lambda:=1) &= \sum_{i=1}^{N} -\attL^{gau}(\hi, \wyi)\\
&= \sum_{i=1}^{N} -\frac{1}{2} \norm{\hi - \wyi}^2
\end{split}
\end{equation}
This is equivalent to the definition of center loss \cite{wen2016discriminative}. When taking the partial derivatives with respect to the weights $\wk$, the closed form solution arises: $\wk = \frac{1}{C_k} \sum_{i=1}^{N} \hi \mathbbm{1}_{k = y_i}$, where $\mathbbm{1}$ is the boolean indicator function and $C_k$ is the number of samples belonging to class $k$. In other words, the COREL formulation results in the conclusion that the $\wk$ should be statically set to be the centroids of the latent representations of samples belonging to class $k$, when $\lambda=1$. This is in accordance with the design choice of center loss, where these $w_k$ are the centroids for the classes of the latent space, which is an upper bound on the best K-means clustering solution in the latent space. However, when $\lambda < 1$, and for the other loss functions, it is suboptimal to statically set the weights to be the centroids, as the centroids are not the solution to the objectives of other loss functions. Thus, the loss can be further minimized by directly learning the weights, rather than statically assigning them.

\section{Experimental Design}
We evaluate our proposed models on image classification and news article classification, using both feed-forward (FFNN) and convolutional neural networks (CNN). Beyond classification accuracy, we evaluate the quality of the representations made by the different loss functions by evaluating the performance of clustering algorithms on the latent spaces.

\subsection{Datasets}
We experiment on the benchmark image classification datasets MNIST \cite{lecun1998gradient} and  Fashion-MNIST \cite{xiao2017fashion}. The latter was recently designed as a more challenging alternative to MNIST. Similarly to MNIST, it contains $28 \times 28$ pixel grayscale images, with 60,000 training set samples (5,000 are split away for validation) and 10,000 test set samples belonging to 10 different object classes. Unlike MNIST, these images are of real physical objects, such as bags, t-shirts, and sneakers. Therefore, the model must be able to distinguish shapes and more varied types of image features more precisely than on MNIST. 

We also experiment on an 8-topic subset of the AGNews dataset\footnote{\url{https://www.di.unipi.it/~gulli/AG_corpus_of_news_articles.html}}, a common benchmark for text classification in NLP \cite{zhang2015character}. We divide the dataset into its 8 largest topics: \textit{U.S.}, \textit{World}, \textit{Europe}, \textit{Sports}, \textit{Science \& Technology}, \textit{Health}, \textit{Business}, and \textit{Entertainment}. We filter each topic to have 12,000 samples, excepting the \textit{Entertainment} topic, which has 10,721 due to additional filtering required to ensure they were truly ``entertainment'' \footnote{See supplemental material for all preprocessing decisions.}. With a total of 94,721 samples, we divide $15\%$ of them into a validation set and $15\%$ into a final test set. Each sample contains the news article's title concatenated with its description. As input to our models, we use the $300$-dimensional 840B Common Crawl pretrained Glove word embeddings \cite{pennington2014glove} provided by Stanford\footnote{\url{https://nlp.stanford.edu/projects/glove/}}.

\subsection{Models}
We experiment on all three datasets with FFNNs and CNNs in order to determine if any COREL variant is more suited to a particular neural architecture. Each experiment uses Adam \cite{kingma2014adam} for stochastic optimization and is trained for 150 epochs with a learning rate of $10^{-4}$ and 128 sample mini-batches. Unless otherwise noted, each layer uses the LeakyReLU activation function with a negative slope of $0.1$ (i.e., $\max(0.1x, x)$). These parameters were tuned during preliminary testing to both optimize the performance of CCE on the datasets and minimize the amount of hyperparameter search necessary for tuning.

On all three datasets, we use FFNNs with two $128$-dimensional fully-connected hidden layers. While the images are flattened to be $784$-dimensional vectors, the word embeddings for text classification are averaged together into one $300$-dimensional input to the FFNN model.

We use a standard CNN architecture for MNIST and Fashion-MNIST. The CNN has three layers of 2D convolutions with filter sizes of 15, 30, and 60, followed by max-pooling with a kernel size of 2. The last activation is followed by two $128$-dimensional fully-connected layers. Our CNN for text classification is inspired by \cite{kim2014convolutional}. As a 1D CNN over word embeddings, it has three convolutional layers with filter sizes of 3, 4, and 5 words, using max-pooling and ReLU activations after each convolution. This model is also followed by two $128$-dimensional fully connected layers. Dropout did not significantly change performance for image classification, but we found it was necessary to include dropout of $0.5$ following the input and between each fully connected layer in the models for text classification.

\subsection{Loss Variants}
We evaluate four different loss variants across each of the six dataset/model combinations. As a baseline, we present results using the default categorical-cross entropy (CCE) loss function, as presented in Equation~\ref{eq:corelcce}. As a more challenging baseline, we also compare to Center Loss \cite{wen2016discriminative}, described below (Equation~\ref{eq:center}). For our presented COREL variants, we experiment with \textit{Cosine}-COREL (Equation~\ref{eq:lcos}) and \textit{Gaussian}-COREL (Equation~\ref{eq:lgaus}), as specific instances of attractive-repulsive loss (Equation~\ref{eq:arl}).

\subsubsection{Center Loss.} Center loss expands upon CCE by using regularization on latent centers in order to improve the discriminative power of deeply learned features. This objective seeks to minimize the squared euclidean distance between latent representations of samples $\hi$ and their class's corresponding center $\bm{\mu}_{y_i}$, the mean vector of all samples belonging to class $y_i$. However, the loss requires additional parameters to be computed -- the dynamically computed latent centers for each class, $\bm{\mu}_1 \ldots \bm{\mu}_K$. The definition of center loss over a mini-batch of $N$ samples is parameterized by some $\lambda \geq 0$:
\begin{equation} \label{eq:center}
	\centerL = \cceL + \frac{\lambda}{2} \sum_{i=1}^{N} \norm{\hi - \bm{\mu}_{y_i}}^{2} 
\end{equation}
It would be computationally infeasible to compute the true latent centers of the entire training set during each update, so an approximation method is used to maintain them, parameterized by a centroid learning rate $\alpha$. During preliminary experimentation we found that varying $\alpha$ did not have a strong impact on performance, so for all experiments we set $\alpha=0.25$. We compare with center loss as it is the most relevant loss function related to clustering that does not require large architectural modifications to neural models.

\subsection{$\lambda$ Tuning}

\begin{figure}[t]
\includegraphics[width=\linewidth]{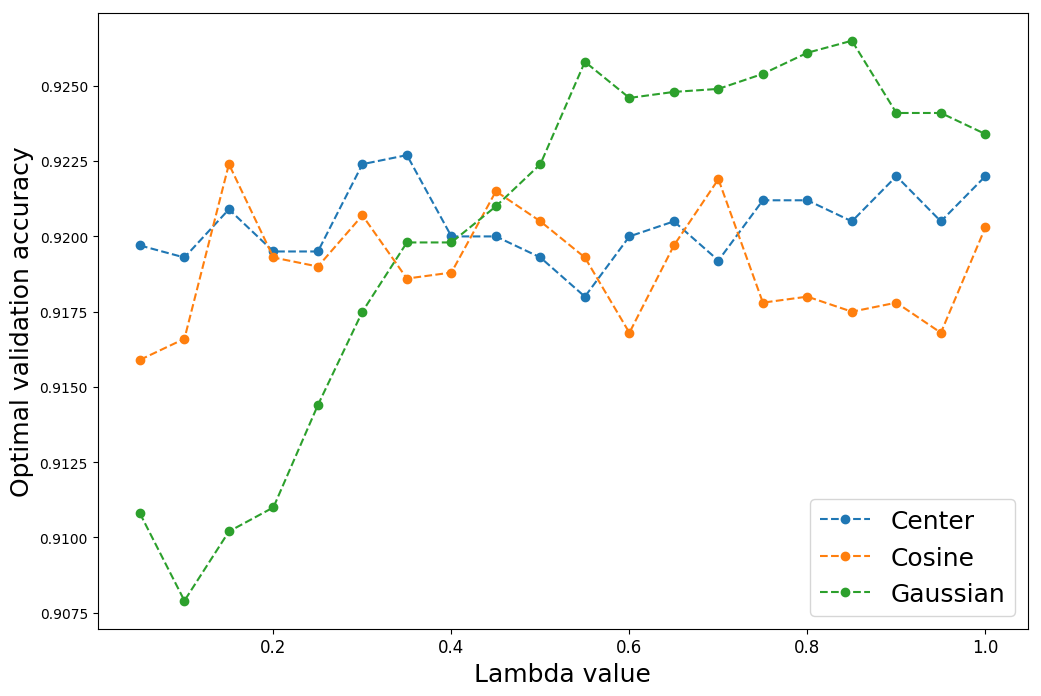}
\caption{$\lambda$ tuning results for CNN on Fashion-MNIST.}
\label{fig:fashionlam}
\end{figure}

Given the six different dataset/model combinations and three different losses that require tuning their $\lambda$ hyperparameters (Center loss, \textit{Cosine}-COREL, \textit{Gaussian}-COREL), we tuned 20 different values of $\lambda$ for each scenario in order to optimize validation set accuracy. The optimally-tuned $\lambda$ values for each setting are presented in Table~\ref{tab:lamtuning}. In Figure~\ref{fig:fashionlam}, we present the specific tuning results for the three losses on Fashion-MNIST with a CNN; figures for all other variants can be found in the supplemental material.

\begin{table}[t]
\centering
	\begin{tabular}{lrrr}
    \toprule
		& \textbf{MNIST} & \textbf{Fashion} & \textbf{AGNews} \\
        \midrule
        Center            & $0.45 / 0.75$ & $0.15 / 0.35$ & $0.20 / 0.25$ \\
        \textit{Cosine}   & $0.20 / 0.50$ & $0.65 / 0.15$ & $0.45 / 0.65$ \\
        \textit{Gaussian} & $0.50 / 0.80$ & $0.50 / 0.85$ & $0.50 / 0.50$ \\
    \bottomrule
	\end{tabular}
\caption{Tuned $\lambda$ values (FFNN/CNN) for each dataset \& model variant from testing $20$ values between $0$ and $1$.}
\label{tab:lamtuning}
\end{table}

The results in Table~\ref{tab:lamtuning} show that shifting the weighting for the attractive and repulsive terms can result in modest improvements over a default $\lambda \smalleq 0.5$. We found that center loss was relatively insensitive to $\lambda$, while \textit{Gaussian}-COREL obtained modest improvements from tuning. 
In the image classification tasks \textit{Cosine}-COREL was relatively insensitive to $\lambda$. However, for text classification there was higher variance -- it was essential for $\lambda$ to weight attraction either evenly with or more than repulsion, otherwise performance was substantially worse. In Figure~\ref{fig:fashionlam} we observe that highly weighting attraction for \textit{Gaussian}-COREL results in substantial gains in validation accuracy over $\lambda\smalleq 0.5$. This suggests that the probabilistic perspective that softmax entails is not necessarily beneficial, as augmenting the weighting by $\lambda\smalleq 0.85$ in this case is better but removes the probabilistic interpretation.

\section{Results}
In Table~\ref{tab:testres} we present the final test set results obtained with each loss variant by averaging their results across ten different runs of the model with different random seeds. The test set accuracy is taken at the epoch where the model obtains the best validation accuracy. Results are presented with standard deviations and are evaluated for statistical significance.

\textit{Gaussian}-COREL is statistically the best in four out of six cases; it is only outperformed when a FFNN is used on MNIST, which is not of particular concern considering that it is better in the standard situation for image classification in which a CNN is used. \textit{Cosine}-COREL has slightly worse performance than the other models, especially for the CNN on text classification and the FFNN on MNIST. However, it does perform slightly better than CCE on MNIST and Fashion-MNIST with CNNs, which is consistent with other work where cosine-based loss functions prove useful for image classification and facial recognition problems \cite{wang2017normface,wang2018cosface}. But in each case \textit{Gaussian}-COREL outperforms \textit{Cosine}-COREL, in terms of classification accuracy.

Center loss has been motivated as making more discriminative features for image classification \cite{wen2016discriminative}. This is consistent with the results for MNIST and Fashion-MNIST, where performance improves upon CCE. However, we observe that  using center loss for regularization actually results in \emph{worse} performance for text classification for both the FFNN and, to a lesser extent, the CNN. Meanwhile, the more discriminative features induced by \textit{Gaussian}-COREL result in improvement over CCE in all six scenarios. This result suggests that more discriminative features are not exclusively useful for image classification and that the \textit{Gaussian}-COREL loss function induces them in a more generalizable way than both center loss and categorical cross-entropy.

\begin{table}[t]
\centering
	\begin{tabular}{lrrr}
    \toprule
		& \textbf{MNIST} & \textbf{Fashion} & \textbf{AGNews} \\
        \midrule
        \multicolumn{4}{l}{\textbf{FFNN}}\\
        CCE               & $97.60 \pm .12$ & $88.31 \pm .18$ & $76.71 \pm .22$ \\
        Center            & $\textbf{97.94} \pm .10$ & $88.68 \pm .21$ & $76.21 \pm .24$ \\
        \textit{Cosine}   & $96.95 \pm .15$ & $88.12 \pm .17$ & $76.81 \pm .22$ \\
        \textit{Gaussian} & $97.66 \pm .08$ & $\underline{88.88} \pm .20$ & $\textbf{77.87} \pm .15$ \\
        \midrule
        \multicolumn{4}{l}{\textbf{CNN}}\\
        CCE               & $99.15 \pm .06$ & $90.95 \pm .24$ & $80.87 \pm .11$ \\
        Center            & $99.39 \pm .05$ & $91.25 \pm .17$ & $80.83 \pm .11$ \\
        \textit{Cosine}   & $99.34 \pm .04$ & $90.99 \pm .20$ & $79.57 \pm .10$ \\
        \textit{Gaussian} & $\underline{99.44} \pm .03$ & $\textbf{91.79} \pm .17$ & $80.90 \pm .13$ \\
    \bottomrule
	\end{tabular}
\caption{Final test set accuracies averaged over 10 runs with different random seeds. Top half is from FFNNs, bottom is CNNs. \underline{Underline} and \textbf{bold} indicate statistically significant improvements upon the next best model with $p < 0.05$ and $p < 0.001$, respectively (paired two-tail \textit{t}-test).}
\label{tab:testres}
\end{table}

\section{Analysis}
Beyond classification accuracy, we analyze the quality of the the representations obtained by using each loss function. In Figure~\ref{fig:viz} we visualize representations obtained by CNNs on MNIST. We first observe that CCE creates a latent space where classes are linearly separable, but not well distinguished. \textit{Cosine}-COREL creates representations where samples of the same classes are highly concentrated along the span of a central weight vector, with minimal angular difference between each other; in this case, the learned weight vectors form a near-orthogonal basis for a $K$-dimensional subspace in the latent space $\h$. \textit{Gaussian}-COREL and center loss seem to make similar latent spaces with well defined-clusters due to the similar form of their loss functions. However, while the two may look similar in a visualization on MNIST, the differences in classification performance for other settings (Table~\ref{tab:testres}) reveals the qualitative differences between the loss functions.



\begin{table}[t]
\centering
  \begin{tabular}{lrrrr}
  \toprule
  \textit{AGNews} & \textbf{Acc} & \textbf{ARI} & \textbf{V-M} & \textbf{Sil.}\\
  \midrule
  \multicolumn{4}{l}{\textbf{K-Means clustering}}\\
	CCE               & $0.692$ & $0.542$ & $0.605$ & $0.266$ \\
    Center            & $0.818$ & $0.645$ & $0.651$ & $0.434$ \\
	\textit{Cosine}   & $0.808$ & $0.626$ & $0.626$ & $0.891$ \\
    \textit{Gaussian} & $0.816$ & $0.641$ & $0.642$ & $0.416$ \\
  \midrule
  \multicolumn{3}{l}{\textbf{Gaussian mixture model}}\\ 
	CCE               & $0.696$ & $0.488$ & $0.583$ & $0.215$ \\
    Center            & $0.718$ & $0.556$ & $0.601$ & $0.373$ \\
	\textit{Cosine}   & $0.809$ & $0.625$ & $0.625$ & $0.891$ \\
    \textit{Gaussian} & $0.675$ & $0.472$ & $0.574$ & $0.347$ \\
  \bottomrule
  \end{tabular}
\caption{Clustering results on test set representations made with a CNN on AGNews for text classification.}
\label{tab:textcluster}
\end{table}

\begin{table}[t]
\centering
  \begin{tabular}{lrrrr}
  \toprule
  \textit{Fashion} & \textbf{Acc} & \textbf{ARI} & \textbf{V-M} & \textbf{Sil.}\\
  \midrule
  \multicolumn{4}{l}{\textbf{K-Means clustering}}\\
	CCE               & $0.729$ & $0.625$ & $0.741$ & $0.299$ \\
    Center            & $0.913$ & $0.824$ & $0.843$ & $0.682$ \\
	\textit{Cosine}   & $0.902$ & $0.803$ & $0.827$ & $0.832$ \\
    \textit{Gaussian} & $0.913$ & $0.824$ & $0.840$ & $0.740$ \\
  \midrule
  \multicolumn{3}{l}{\textbf{Gaussian mixture model}}\\ 
	CCE               & $0.731$ & $0.628$ & $0.744$ & $0.296$ \\
    Center            & $0.909$ & $0.816$ & $0.839$ & $0.668$ \\
	\textit{Cosine}   & $0.902$ & $0.803$ & $0.827$ & $0.832$ \\
    \textit{Gaussian} & $0.913$ & $0.824$ & $0.840$ & $0.740$ \\
  \bottomrule
  \end{tabular}
\caption{Clustering results on test set representations made with a CNN on Fashion-MNIST for image classification.}
\label{tab:fashioncluster}
\end{table}

Next, to evaluate the clusterability of the latent spaces made by our models, we present results from applying parametric clustering algorithms (K-Means and Gaussian Mixture model) on the latent spaces made by the different loss variants on AGNews (Table~\ref{tab:textcluster}) and Fashion-MNIST (Table~\ref{tab:fashioncluster})\footnote{The accuracy (\textbf{Acc}) evaluates clustering performance by using the Hungarian algorithm to align cluster predictions with the most likely class predictions \cite{munkres1957algorithms}; the adjusted random index (\textbf{ARI}) computes how far the cluster predictions are from a random clustering \cite{vinh2010information}; the V-Measure (\textbf{V-M}) is an entropy-based metric that computes harmonic mean of homogeneity and completeness \cite{rosenberg2007v}; these three are supervised metrics that compare the cluster predictions with the ground truth labels.  The Silhouette coefficient (\textbf{Sil.}) is an unsupervised metric that measures the intra- versus inter-class variance between representations and cluster predictions; i.e., cluster density versus cluster separability \cite{rousseeuw1987silhouettes}.}.

We first observe that the representations made by CCE are not naturally clusterable; for AGNews (Table~\ref{tab:textcluster}) performance degrades by over $11\%$ in accuracy from its original accuracy of $80.87$ (Table~\ref{tab:testres}), while it degrades by over $17\%$ for Fashion-MNIST (Table~\ref{tab:fashioncluster}). This shows that the linear separability induced by CCE is not equivalent to making a naturally clusterable latent space. However, the results from including center loss (Equation~\ref{eq:center}) with the CCE objective can make the latent space significantly more clusterable.

There are qualitative differences between the representations made for the different problem domains, in terms of clusterability. While center loss and \textit{Gaussian}-COREL make naturally clusterable representations for both K-Means and a Gaussian Mixture model (GM) on images, they are more difficult to cluster with a GM for text classification. \textit{Cosine}-COREL, however, makes generally naturally clusterable representations, regardless of the clustering algorithm or domain, as evidenced by the consistency in accuracy, ARI, and V-measure scores. Moreover, we observe that \textit{Cosine}-COREL makes clusters that are more well-defined and internally consistent, as evidenced by its high Silhouette scores when compared with the other models.

These results show that \textit{Gaussian}-COREL and Center loss can make naturally clusterable representations, but are more sensitive to the clustering algorithm, depending on the problem domain. In general, \textit{Cosine}-COREL makes naturally clusterable representations with more well-defined clusters than the other models.

\section{Related Work}
Recently there have been many approaches using either cosine- or Gaussian-based loss functions. Most of these are used explicitly for the domain of image classification, where the problem of needing discriminative features is only understood as necessary for image problems, particularly facial recognition \cite{wang2017normface,ranjan2017l2,gao2018margin,wang2018cosface,zheng2018ring}. Some recent work has combined cosine similarity with weight imprinting \cite{qi2017learning}, which sets $\W$ for $\pred$ to be dynamically computed latent centroids (as in center loss \cite{wen2016discriminative}); they then apply the softmax operation over the cosine similarities, as in congenerous cosine loss \cite{liu2017learning}. Other work models image classes as Gaussians \cite{wan2018rethinking}, but requires learning the covariance matrix $\Sigma$ in the similarity function, which is constrained to be diagonal. 

In natural language processing, cosine similarity-based losses have only begun to be explored for the purpose of constructing more meaningful representations. In one case for the purpose of linearly constructing antonymous word embeddings \cite{mrkvsic2016counter}, and also in a deep transfer learning task of building clusterable event representations for event coreference resolution \cite{kenyon2018resolving}. 

Recent work \cite{rippel2015metric} has modelled classes with a set of Gaussians with the motivation of creating a well-structured latent space, using neighborhood-based sampling to maintain the centers of these Gaussians. However, this requires substantial architectural modifications to neural models, requiring frequent pauses during training to run a K-Means clustering algorithm over the latent space, becoming more costly as the training set size increases. Other work has designed similarly motivated loss functions, but also requires significantly more model engineering than COREL. This includes pairwise-based methods \cite{hsu2015neural,fogel2018clustering,hsu2018learning}, and triplet-based methods \cite{norouzi2012hamming,schroff2015facenet,he2018triplet}; all of these require sophisticated methods for sampling training data, necessitating more hyperparameters and architectural modifications in order to implement their methods. 

In this work, we compared to center loss (Equation~\ref{eq:center}) as the most related loss function to our COREL models. As we explained, other loss functions exist that are motivated toward constructing more discriminative features for specific image classification and facial recognition problems, but either require substantial model engineering or are not motivated toward the problem of natural clusterability. 

\section{Conclusion}
We have proposed \textit{clustering-oriented representation learning} (COREL) which optimizes a new general class of optimization objectives which we call \textit{attractive-repulsive} (AR) loss. AR loss is a general framework we propose for understanding and designing loss functions in neural networks in terms of some \textit{similarity function}. We showed that categorical cross-entropy (CCE) is a specific instance of AR loss where the similarity function is defined as the inner product. We proposed two variants of COREL based on the intuitions garnered from our formulation of AR loss. The first is \textit{Cosine}-COREL, which uses cosine similarity in its loss function and is not reliant on the softmax operation. The second variant is \textit{Gaussian}-COREL, which uses the softmax operation with the negative squared euclidean distance similarity function.

We presented experiments on three different datasets, covering both image and text classification. We compared CCE and center loss \cite{wen2016discriminative} with our proposed \textit{Cosine}-COREL and \textit{Gaussian}-COREL loss functions. The results showed that \textit{Gaussian}-COREL reliably builds representations that outperform both CCE and center loss in terms of classification accuracy. 
We evaluated the natural clusterability of the latent spaces by applying standard clustering algorithms on the representations produced at the final hidden layers of our trained neural models. While \textit{Cosine}-COREL does not do as well as \textit{Gaussian}-COREL in terms of accuracy, it consistently produces a latent space that is naturally clusterable into its object classes, regardless of the problem domain.
 
In future work, we would like to determine any mathematical properties that attractive and repulsive losses need to possess with respect to the optimization loss surface. We believe that naturally clusterable representations could prove useful in problem settings where representations made by supervised models are used in downstream tasks, and seek to evaluate our COREL variants in these situations.

\section*{Acknowledgements}
This work was funded in part by the Fonds de recherche du Qu\'{e}bec (FRQNT), the Natural Sciences and Engineering Research Council (NSERC), and Compute Canada.

\bibliography{aaai}
\bibliographystyle{aaaai}

\end{document}